\documentclass[11pt,reqno]{article}
\usepackage{jmlr2e}
\usepackage{fullpage,times,graphicx,amssymb,amsmath,amsfonts,bbm,psfrag,xcolor}

\usepackage{graphicx}
\usepackage{amssymb}
\usepackage{amsmath}
\usepackage{amsfonts}
\usepackage{bbm}
\usepackage{psfrag}
\usepackage{xcolor}
\usepackage{fullpage}
\usepackage{epstopdf}
\usepackage{graphicx}
\usepackage{times}
\usepackage{jmlr2e}
\usepackage{commath}
\usepackage{thmtools}
\usepackage{thm-restate}
\usepackage{times}
\usepackage{graphicx}
\usepackage{natbib}
\usepackage{dirtytalk}
\definecolor{ddarkbrown}{rgb}{0.5,0.2,0.05} \definecolor{bbluegray}{rgb}{0.05,0,0.5}

\usepackage[colorlinks,citecolor=bbluegray,linkcolor=ddarkbrown,urlcolor=blue,breaklinks]{hyperref}

\usepackage{subfig,float} 

\usepackage{mathtools}

\usepackage{algorithm,algcompatible,algpseudocode}
\usepackage{multicol,lipsum}
\algnewcommand{\Inputs}[1]{%
	\State \textbf{Inputs: \:}{#1}
}

\algnewcommand{\Output}[1]{%
	\State \textbf{Output: \:}{#1}
}
\algnewcommand{\Initialize}[1]{%
	\State \textbf{Initialize: \:}{#1}
}

\algnewcommand{\IIf}[1]{\State\algorithmicif\ #1\ \algorithmicthen}
\algnewcommand{\EndIIf}{\unskip\ \algorithmicend\ \algorithmicif}

\let \oldsection \section
\renewcommand{\section}{\vspace{3ex plus 1ex}\oldsection}

\newcommand{\BEAS}{\begin{eqnarray*}}
	\newcommand{\EEAS}{\end{eqnarray*}}
\newcommand{\BEA}{\begin{eqnarray}}
\newcommand{\EEA}{\end{eqnarray}}
\newcommand{\mb}{\mathbb}
\newcommand{\BEQ}{\begin{equation}}
\newcommand{\EEQ}{\end{equation}}
\newcommand{\BIT}{\begin{itemize}}
	\newcommand{\EIT}{\end{itemize}}
\newcommand{\BNUM}{\begin{enumerate}}
	\newcommand{\ENUM}{\end{enumerate}}

	\newcommand{\D}{\mathcal{D}}
	\newcommand{\Sk}{\mathcal{S}}
		\newcommand{\U}{\mathcal{U}}
	\newcommand{\Pm}{\mathcal{P}}
	\newcommand{\te}{\theta}
	
	\newcommand{\R}{\mathbb{R}}

\newcommand{\BA}{\begin{array}}
	\newcommand{\EA}{\end{array}}

 \numberwithin{dummy}{section}

\numberwithin{mythm}{section}
\numberwithin{mydef}{section}
\numberwithin{myprop}{section}
\numberwithin{mylem}{section}
\numberwithin{mycor}{section}

\title{Operator Sketching for Deep Unrolling Networks}

\begin{document}
	\author{Junqi Tang, Subhadip Mukherjee, Carola-Bibiane Sch\"onlieb \email {jt814, sm2467, cbs31}@cam.ac.uk\\
		\addr Department of Applied Mathematics and Theoretical Physics (DAMTP),\\ University of Cambridge
	}
	\editor{}
	
	
	\maketitle

\begin{abstract}

    In this work we propose a new paradigm for designing efficient deep unrolling networks using operator sketching. The deep unrolling networks are currently the state-of-the-art solutions for imaging inverse problems. However, for high-dimensional imaging tasks, especially the 3D cone-beam X-ray CT and 4D MRI imaging, the deep unrolling schemes typically become inefficient both in terms of memory and computation, due to the need of computing multiple times the high-dimensional forward and adjoint operators. Recently researchers have found that such limitations can be partially addressed by stochastic unrolling with subsets of operators, inspired by the success of stochastic first-order optimization. In this work, we propose a further acceleration upon stochastic unrolling, using sketching techniques to approximate products in the high-dimensional image space. The operator sketching can be jointly applied with stochastic unrolling for the best acceleration and compression performance. Our numerical experiments on X-ray CT image reconstruction demonstrate the remarkable effectiveness of our sketched unrolling schemes.

\end{abstract}

\section{Introduction}

Deep-learning solutions have started their dominance in computational imaging applications with remarkable state-of-the-art performance \citep{zhang2017beyond,jin2017deep}. In particular, the deep-unrolling networks \citep{gregor2010learning,adler2018learned,xiang2021fista,tang2021stochastic}, which are based on unfolding the iterations of classical proximal-splitting methods \citep{combettes2011proximal}, such as ISTA/FISTA \citep{beck2009fast}, ADMM \citep{boyd2011distributed}, PDHG \citep{chambolle2011first}, or stochastic PDHG \citep{chambolle2018stochastic}, explicitly encode the measurement physics within the network and achieve the state-of-the-art performance and stability in solving imaging inverse problems. Among these deep unrolling networks, the Learned Primal-Dual (LPD) network of \cite{adler2018learned} is the most popular and widely applied due to its excellent effectiveness and expressibility -- it is a general network framework which covers most of other unrolling networks such as Learned ISTA/FISTA as special cases.

Despite the impressive performance, there is a major limitation in deep unrolling schemes, and also the related plug-and-play (PnP) schemes with pretrained denoising networks \citep{venkatakrishnan2013plug,kamilov2017plug}. The unrolling networks typically need to compute multiple times the forward and adjoint operators in a single forward pass. For large-scale high-dimensional imaging tasks, for instance 3D X-ray CT and 4D MRI reconstruction, such networks become very difficult to train since it will require huge memory and computational effort, and computationally inefficient in testing as well.

To mitigate the computational cost of PnP methods, stochastic gradient/Hessian estimators have been used to accclerate full-batch PnP schemes \citep{sun2019online,liu2020rare,perelli2021regularization,tang2020fast,sun2021scalable}. Most recently, \cite{tang2021stochastic} proposed a stochastic extension of LPD, namely the Learned Stochastic Primal-Dual (LSPD), to mitigate this computational bottleneck of deep unrolling schemes.

Very recently, \cite{tang2022accelerating} proposed to further accelerate both full-batch and stochasic PnP methods with multi-stage sketched gradient updates. Unlike other existing sketching techniques which performs random projections in data domain \citep{pilanci2017newton,2016_Pilanci_Iterative,pmlr-v70-tang17a}, this sketching of \cite{tang2022accelerating} is performed in the image domain with downsampling interpolation, and approximate the high-dimensional full gradient by upsampling the sketched gradient from the sketched low-dimensional image space, hence a significant computational gain can be obtained. In this work we extend this idea of sketched gradient to deep unrolling networks by proposing the operator sketching to provide further acceleration of LPD/LSPD.

\section{Background}
Imaging inverse problems, such as image deblurring/inpainting/superresolution, and CT/MRI/PET tomographic image reconstruction can be generally expressed as:
\begin{equation}\label{model}
   b = A x^\dagger + w,
\end{equation}
where $x^\dagger \in \mb{R}^d$ is the ground truth image, $A \in \mb{R}^{n \times d}$ is the forward operator,, $b \in \mb{R}^n$ denotes the measurement data, and $w \in \mb{R}^n$ the measurement noise (can be data-dependent).

In order to obtain a good estimation of the ground-truth image $x^\dagger$, a classical approach is to solve a composite optimization problem which contains a data-fitting term $f(b, Ax)$ enforcing consistency towards the measurement data we got, and a regularization term $r(x)$ modeling the prior knowledge of the image:
\begin{equation}\label{1st-stage}
    x^\star \in \arg\min_{x \in \mb{R}^d} f(b, Ax) + r(x).
\end{equation}
Practitioners typically choose $f(\cdot)$ to be some smooth convex function such as the least-squares loss, while the classical choices of $r(\cdot)$ includes the $\ell_1$ penalty on wavelet/shearlet domain enforcing sparsity on such transformed domain, and the total-variation (TV) regularizer enforcing sparsity on the image gradients.

The composite optimization problem (\ref{1st-stage}) can be efficiently solved by proximal splitting algorithms \citep{combettes2011proximal} such as the primal-dual hybrid gradient (PDHG) of \cite{chambolle2011first}. The PDHG finds the minimizer of (\ref{1st-stage}) by solving the saddle-point problem which is the primal-dual reformulation of (\ref{1st-stage}):

\begin{equation}\label{saddle}
    [x^\star, y^\star] = \min_{x} \max_{y} \{r(x) + \langle Ax, y \rangle - f_b^*(y)\},
\end{equation}
where $f_b^*(\cdot)$ is defined as:
\begin{equation}
    f_b^*(y) := \sup_{h} \{\langle h, y \rangle - f_b(h)\},
\end{equation}
which is known as the Fenchel conjugate.

The PDHG algorithm of \cite{chambolle2011first} can be written as the following:

 \begin{eqnarray*}
 && \mathrm{\textbf{Primal-Dual Hybrid Gradient (PDHG) of \cite{chambolle2011first}}} \\&& - \mathrm{Initialize}\ x_0, \Bar{x}_0 \in \mb{R}^d \ y_0 \in \mb{R}^p\\
 &&\mathrm{For} \ \ \ k = 0, 1, 2,...,  K\\
&&\left\lfloor
\begin{array}{l}
y_{k+1} = \mathrm{prox}_{\sigma f_b^*} (y_k + \sigma A \Bar{x}_k);\\
x_{k+1} = \mathrm{prox}_{\tau r} (x_k - \tau A^T y_{k+1});\\
\Bar{x}_{k + 1} = x_{k+1} + \beta_k (x_{k+1} - x_{k});
\end{array}
\right.
 \end{eqnarray*}
where it alternate between dual proximal gradient ascent and primal proximal gradient descent with a momentum step for acceleration. One can exploit the finite-sum structure of some problems (for example CT/PET) for further acceleration. For such problems the saddle-point formulation can be written as:
 \begin{equation}\label{saddle1}
    [x^\star, y^\star] = \min_{x} \max_{y} \{r(x) + \sum_{i = 0}^{m-1}\langle M_iAx, y_i \rangle - f_{b_i}^*(y_i)\}.
\end{equation}
where $\mathbf{M}:= [M_1, M_2, ..., M_m]$ denotes the set of subsampling operators. Then the objective (\ref{saddle1}) can be efficiently solved by stochastic PDHG (SPDHG) of \cite{chambolle2018stochastic}. The Learned Primal-Dual (LPD) network of \cite{adler2018learned} and the Learned Stochastic Primal-Dual (LSPD) network of \cite{tang2021stochastic} are based on unfolding the iterations of PDHG and SPDHG by replacing the proximal operators by convolutional neural networks and trained end-to-end.

 \section{Accelerating the Deep Unrolling Schemes via Operator Sketching}
 In this section we start by introducting the LPD and LSPD networks and then present our sketched acceleration of them. The LPD replaces the proximal operators in the dual and primal steps by the convolutional subnetworks $\D_{\te_d^k}$ and $\Pm_{\te_p^k}$ which are trainable w.r.t. the parameter sets $\te_d^k$ and $\te_p^k$ (meanwhile the step sizes $\tau_k$ and $\sigma_k$ are also trainable parameters), respectively:
 
 \begin{eqnarray*}
 && \mathrm{\textbf{Learned Primal-Dual (LPD) of \cite{adler2018learned}}} \\&&- \mathrm{Initialize}\ x_0 \in \mb{R}^d \ y_0 \in \mb{R}^p\\
 &&\mathrm{For} \ \ \ k = 0, 1, 2,...,  K-1\\
&&\left\lfloor
\begin{array}{l}
y_{k+1} = \D_{\te_d^k} (y_k, \sigma_k, A {x}_k , b);\\
x_{k+1} = \Pm_{\te_p^k}(x_k, \tau_k, A^T y_{k+1});\\
\end{array}
\right.
 \end{eqnarray*}
 
The LPD is a very generic unrolling framework which contains most of exisiting unrolling schemes such as Learned ISTA/FISTA as special cases, hence we focus our study on LPD but the same principle can be applied in any other unrolling network. The LSPD network of \cite{tang2021stochastic} improves the computational and memory efficiency of LPD by minibatching the forward and adjoint operators in each layer:

 \begin{eqnarray*}
 && \mathrm{\textbf{Learned Stochastic Primal-Dual (LSPD) of \cite{tang2021stochastic}}} \\&&- \mathrm{Initialize}\ x_0 \in \mb{R}^d \ y_0 \in \mb{R}^{p/m}\\
 &&\mathrm{For} \ \ \ k = 0, 1, 2,...,  K-1\\
&&\left\lfloor
\begin{array}{l}
i = \mod(k,m); \\\mathrm{(or\ pick\ i\ from\ [0, m-1]\ uniformly\ at\ random)}\\
y_{k+1} = \D_{\te_d^k} (y_k, \sigma_k, (M_iA) {x}_k , M_ib);\\
x_{k+1} = \Pm_{\te_p^k}(x_k, \tau_k, (M_iA)^T y_{k+1});\\
\end{array}
\right.
 \end{eqnarray*}

where $\mathbf{M}:= [M_1, M_2, ..., M_m]$ denotes the set of subsampling operators. For LPD and LSPD (and also the sketched versions of them), we can also consider an optional momentum acceleration with the memory of past $P$ algorithmic layers: 
\begin{equation}\label{pm}
    x_{k+1} = \Pm_{\te_p^k}(X_k, \tau_k, (M_iA)^T y_{k+1}),
\end{equation}
 where $X_k = [x_k, x_{k-1},.. x_{k-P}]$ and also the dual momentum 
 \begin{equation}\label{dm}
     y_{k+1} = \D_{\te_d^k} (Y_k, \sigma_k, (M_iA) {x}_k , M_ib)
 \end{equation}
 where $Y_k = [y_k, y_{k-1},.. y_{k-P}]$, at the cost of extra computation and memory storage. The momentum memory $P$ is recommended to be small in large-scale and high-dimensional tasks for storage efficiency.
 
 Now we are ready to present our sketched LPD and sketched LSPD networks. Our main idea is to speedily approximate the products $A {x}_k$, $A^T y_{k+1}$:
 \begin{equation}\label{skaf}
     A {x}_k \approx A_{s_k} \Sk_{\te_s^k}({x}_k), \ \ A^T y_{k+1} \approx \U_{\te_u^k}(A_{s_k}^T y_{k+1})
 \end{equation}
  where $\Sk_{\te_s^k}(\cdot): \R^d \rightarrow \R^{d_{s_k}}$ ($d_{s_k} < d$) being the sketching/downsampling operator which can be potentially trainable w.r.t parameters $\te_s^k$, while $A_{s_k} \in \R^{n \times d_{s_k}}$ is the sketched forward operator discretized on the reduced low-dimensional image space, and for the dual step we have $\U_{\te_u^k} : \R^{d_{s_k}} \rightarrow \R^d$ the upsampling operator which can also be trained. In practice, we found that it is actually suffice for us to just use the most simple off-the-shelf up/down-sampling operators in Pytorch for example the bilinear interpolation to deliver excellent performance for the sketched unrolling networks. Our Sketched LPD network is written as:
\begin{eqnarray*}
 && \mathrm{\textbf{Sketched LPD} (Option 1)} - \mathrm{Initialize}\ x_0 \in \mb{R}^d \ y_0 \in \mb{R}^p\\
 &&\mathrm{For} \ \ \ k = 0, 1, 2,...,  K-1\\
&&\left\lfloor
\begin{array}{l}
y_{k+1} = \D_{\te_d^k} (y_k, \sigma_k, A_{s_k} \Sk_{\te_s^k}({x}_k) , b);\\
x_{k+1} = \Pm_{\te_p^k}(x_k, \tau_k, \U_{\te_u^k}(A_{s_k}^T y_{k+1}));\\
\end{array}
\right.
 \end{eqnarray*}
Or alternatively:
\begin{eqnarray*}
 && \mathrm{\textbf{Sketched LPD} (Option 2)} - \mathrm{Initialize}\ x_0 \in \mb{R}^d \ y_0 \in \mb{R}^p\\
 &&\mathrm{For} \ \ \ k = 0, 1, 2,...,  K-1\\
&&\left\lfloor
\begin{array}{l}
y_{k+1} = \D_{\te_d^k} (y_k, \sigma_k, A_{s_k} \Sk_{\te_s^k}({x}_k) , b);\\
x_{k+1} = \U_{\te_u^k}(\Pm_{\te_p^k}(\Sk_{\te_s^k}({x}_k), \tau_k, A_{s_k}^T y_{k+1}));\\
\end{array}
\right.
 \end{eqnarray*}
where the sketch size is suggested to be chosen in a ``coarse-to-fine'' manner (more aggressive sketch at the beginning for efficiency, and conservative sketch or non-sketch at latter iterations for accuracy) and the difference between the two option is the ordering of the subnetworks and image up/down samplers.

Again, we can use the same approximation for stochastic gradient steps:
 \begin{equation}\label{ska}
 \begin{split}
     &(M_i A) {x}_k \approx (M_iA_{s_k}) \Sk_{\te_s^k}({x}_k),\\ & (M_i A)^T y_{k+1} \approx \U_{\te_u^k}((M_iA_{s_k})^T y_{k+1}),
    \end{split}
 \end{equation}
and hence we can write our Sketched LSPD network as:
  \begin{eqnarray*}
 && \mathrm{\textbf{Sketched LSPD}(Option 1)} \\&&- \mathrm{Initialize}\ x_0 \in \mb{R}^d \ y_0 \in \mb{R}^{p/m}\\
 &&\mathrm{For} \ \ \ k = 0, 1, 2,...,  K-1\\
&&\left\lfloor
\begin{array}{l}
i = \mod(k,m); \\\mathrm{(or\ pick\ i\ from\ [0, m-1]\ uniformly\ at\ random)}\\
y_{k+1} = \D_{\te_d^k} (y_k, \sigma_k, (M_iA_{s_k}) \Sk_{\te_s^k}({x}_k) , M_ib);\\
x_{k+1} = \Pm_{\te_p^k}(x_k, \tau_k, \U_{\te_u^k}((M_iA_{s_k})^T y_{k+1}));\\
\end{array}
\right.
 \end{eqnarray*}
or alternatively:
  \begin{eqnarray*}
 && \mathrm{\textbf{Sketched LSPD} (Option 2)} \\&&- \mathrm{Initialize}\ x_0 \in \mb{R}^d \ y_0 \in \mb{R}^{p/m}\\
 &&\mathrm{For} \ \ \ k = 0, 1, 2,...,  K-1\\
&&\left\lfloor
\begin{array}{l}
i = \mod(k,m); \\\mathrm{(or\ pick\ i\ from\ [0, m-1]\ uniformly\ at\ random)}\\
y_{k+1} = \D_{\te_d^k} (y_k, \sigma_k, (M_iA_{s_k}) \Sk_{\te_s^k}({x}_k) , M_ib);\\
x_{k+1} = \U_{\te_u^k}(\Pm_{\te_p^k}(\Sk_{\te_s^k}(x_k), \tau_k, (M_iA_{s_k})^T y_{k+1}));\\
\end{array}
\right.
 \end{eqnarray*}
 
 \textbf{Remark regarding varying ``coarse-to-fine'' sketch size for SkLPD and SkLSPD.} Numerically we suggest that we should use more aggressive sketch at the beginning for efficiency, while conservative sketch or non-sketch at latter iterations for accuracy. One plausible choice we found numerically pretty successful is: for the last few unrolling layers of SkLPD and SkLSPD, we switch to usual LPD/LSPD (say if the number of unrolling layers is 20, we can choose last 4 unrolling layers to be unsketched, that is, $A_{s_k} = A$ for $k>K_{\mathrm{switch}}$), such that the reconstruction accuracy is best preserved.
 
 \textbf{Remark regarding the Option 2 for further improving efficiency.} The second option of our SkLPD and SkLSPD further accelerates the computation comparing to Option 1, by making the primal-subnet taking the low-dimensional images and gradients as input and then upscale.  Noting that the usual choice for the up and down sampler would simply be an off-the-shelf interpolation algorithm such as bilinear or bicubic interpolation which can be very efficiently computed, in practice we found the optional 2 often more favorable computationally if we use the coarse-to-fine sketch size. Numerically we found SkLPD and SkLSPD with option 2 and coarse-to-fine sketch size can be both trained faster and more efficient in testing due to the further reduction on the computation of the primal-subnet, without loss on reconstruction accuracy comparing to option 1.

Note that for the Sketched LPD/LSPD we can also use the optional learned momentum presented in equations (\ref{pm}) and (\ref{dm}) with an extra cost on computation and storage.

\begin{table*}[t]\label{Syn}
\caption{Sparse-view fan-beam CT Result for sketched LPD and LSPD on Mayo dataset}
\label{sample-table1}
\vskip 0.15in
\begin{center}
\begin{small}
\begin{sc}
\begin{tabular}{lcccr}
\hline
Method & $\#$ calls on $A$ and $A^T$& PSNR & SSIM  \\
\hline

&&&\\

LPD \textit{(12 layers)} & 24& 39.3483 & 0.9398  \\

LSPD  \textit{(12 layers)} & 6& 38.1044& 0.9162\\
\\

Sketched LPD \textit{(12 layers)} & 12& 38.5842 & 0.9289  \\

Sketched LSPD  \textit{(12 layers)} & 3& 37.6573& 0.9260\\

\hline
\end{tabular}
\end{sc}
\end{small}
\end{center}
\vskip -0.1in
\end{table*}

 \begin{figure}[t]
   \centering

    {\includegraphics[trim=80 50 15
    25,clip,width=0.9\textwidth]{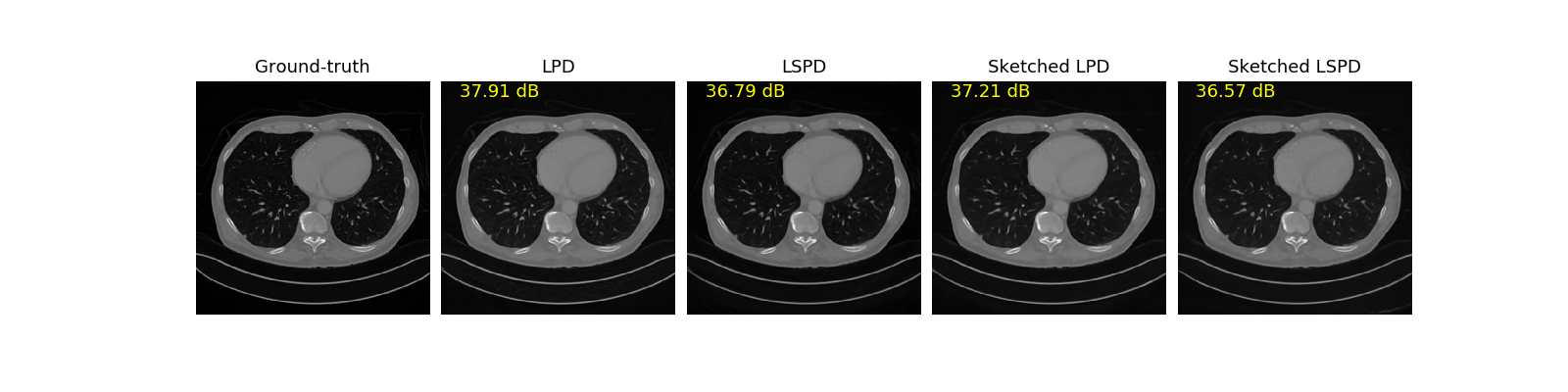}}
   {\includegraphics[trim=80 50 15
    55,clip,width=0.9\textwidth]{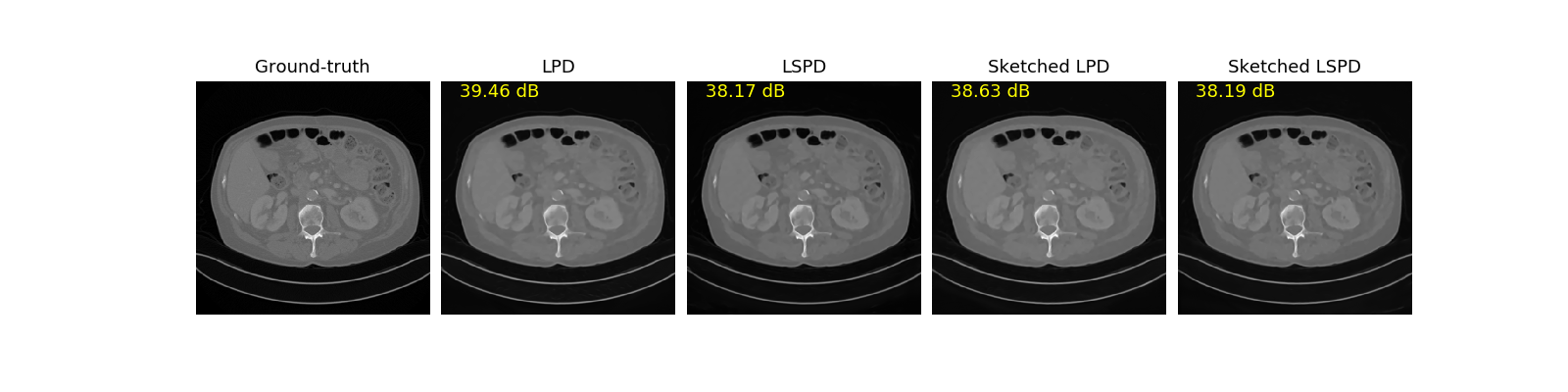}}
      {\includegraphics[trim=80 50 15 55,clip,width=0.9\textwidth]{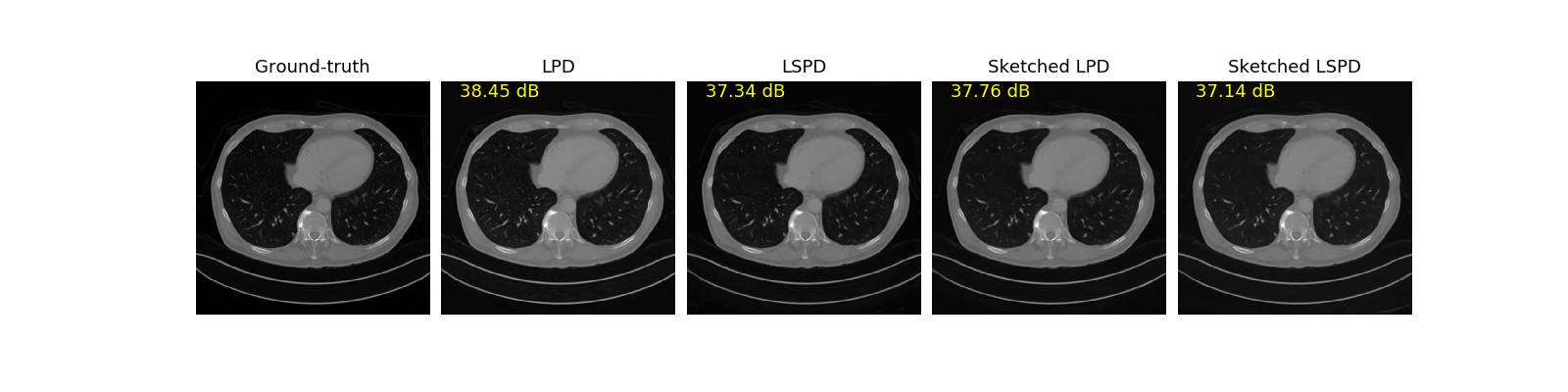}}
         {\includegraphics[trim=80 50 15
    55,clip,width=0.9\textwidth]{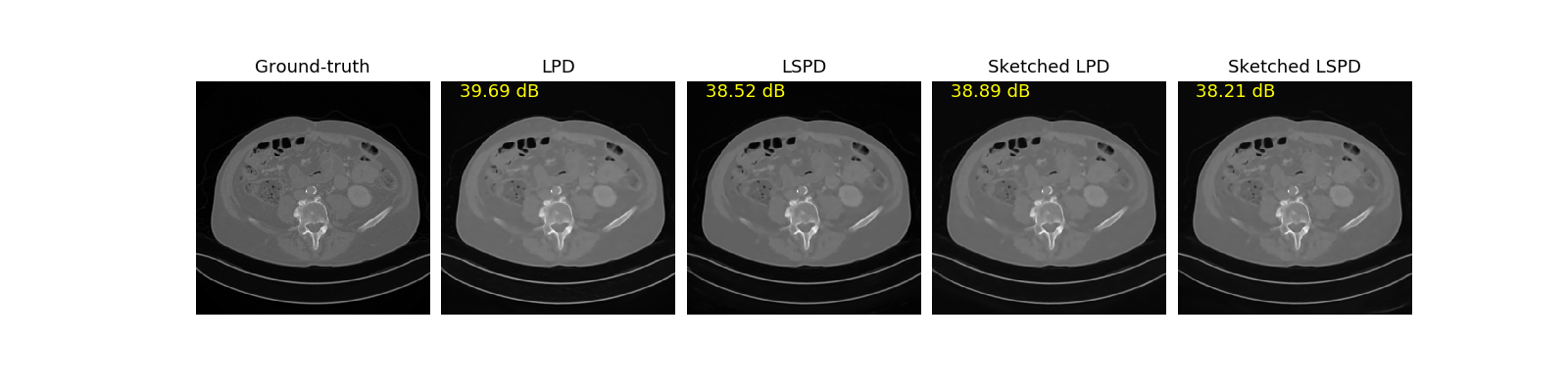}}
            {\includegraphics[trim=80 50 15
    55,clip,width=0.9\textwidth]{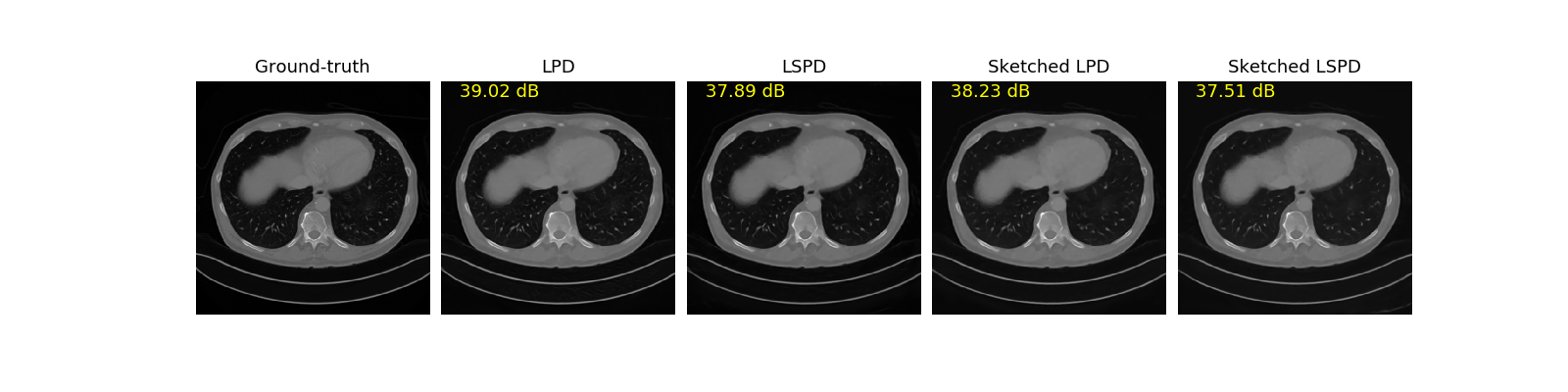}}

   \caption{Examples of the testing results}
   \label{f2}
\end{figure}

\section{Numerical Experiments}

In this section we present our numerical experiments on sparse-view fan-beam X-ray CT image reconstruction. Our setting here is exactly the same as the sparse-view CT experiment presented in the Appendix E of \citep{tang2021stochastic}. We use the Clinic-Mayo dataset \citep{mccollough2016tu} containing 10 patients for training and testing. We use 2111 slices from 9 patients for training and 118 slices from 1 patient for testing, each slice is sized 512 by 512. We simulate using ODL toolbox \citep{adler2018learned} the fan beam projection data with 200 equally-spaced projections and each one of it includes 400 rays for measurements, which are corrupted with poisson noise:

\begin{equation}
    b \sim \mathrm{Poisson}(I_0 e^{-Ax^\dagger}),
\end{equation}
with $I_0 = 5 \times 10^6$ and we take the logarithmic to linearize the measurements.

We compare our Sketched LPD and Sketched LSPD with original LPD and LSPD. For all these networks, we choose the subnetworks $\Pm_{\te_k}$ and $\D_{\te_k}$ to have 3 convolutional layers (with a skip connection between the first channel of input and the output) and 32 channels, with kernel size 5, and we do not use momentum option for simplicity and memory efficiency. The starting point $x_0$ is set to be the standard filtered-backprojection for all the unrolling networks. We set all of them to have 12 algorithmic layers ($K=12$), while for LSPD we partition the forward/adjoint operators into 4 subsets. For the up/down-samplers in our Sketched LPD and Sketched LSPD, we simply choose the bilinear upsample and downsample functions in Pytorch. When called, the up-sampler increase the input image 4 times larger (from $256 \times 256$ to $512 \times 512$), while the down-sampler makes the input image 4 times smaller (from $512 \times 512$ to $256 \times 256$). While the full forward operator $A$ is defined on the grid of $512 \times 512$, the sketched operator $A_s$ \footnote{Here we choose to make $A_{s_k}$ to stay the same for all $k$. We may consider also a varying coarse-to-fine sketch size for even better compression as proposed in Section 3. We restrain from reporting these results here at the moment but include the detailed numerical study in the journal version.} is defined on the grid of $256 \times 256$ hence requires only a half of the computation in this setting. We train all the networks with supervised end-to-end learning, using 20 epochs of Adam \citep{kingma2014adam}.

We present our testing results in Table 1 and Figure 1. We can observe that our sketched unrolling networks achieve very similar performance in PSNR and SSIM with unsketched counterparts, with only a fraction of computation. There is no visible difference in the reconstructed images between sketched and unsketched networks. Particularly, our Sketched LSPD achieves the best compression on the computational cost on the forward/adjoint operators, showing that the best efficiency can be achieved by jointly utilizing the operator sketching and stochastic gradient approximation.

\section{Conclusion}

In this paper we propose a new paradigm for the acceleration and compression of the deep unrolling network, using sketching techniques to perform dimensionality reduction. We consider the LPD of \cite{adler2018learned} and LSPD of \cite{tang2021stochastic} networks as the candidate of our sketched acceleration since these are very generic unrolling framework, including most of existing unrolling networks as special cases. A theoretical analysis of estimation error of Sketched LPD/LSPD can be done with similar steps in the proof of \citep[Theorem 3.1]{tang2021stochastic}, with an additional assumption on the boundedness of the sketching approximation errors in (\ref{skaf}) and (\ref{ska}). Our numerical results on sparse-view CT image reconstruction demonstrate that by jointly utilize the operator sketching and stochastic gradient estimation we can achieve state-of-the-art acceleration and compression of deep unrolling networks, almost without any compromise on reconstruction accuracy.

\bibliography{main.bib}

\end{document}